\documentclass[sigconf,pbalance]{acmart}

\usepackage{amsmath}
\usepackage{booktabs}
\usepackage{multirow}
\usepackage{graphicx}
\usepackage{xcolor}
\usepackage{colortbl}
\usepackage{algorithm}
\usepackage{algpseudocode}
\usepackage{microtype}
\usepackage{xspace}
\definecolor{mosaiclight}{HTML}{EAF1FA}
\definecolor{mosaicblue}{HTML}{315F8C}
\newcommand{\method}{\textsc{FedRCA}\xspace}
\newcommand{\acc}[2]{#1{\scriptsize$\pm$#2}}
\newcommand{\best}[1]{\textbf{#1}}
\newcommand{\second}[1]{\underline{#1}}
\newcommand{\stabilized}[1]{\textcolor{gray}{#1}}
\algrenewcommand\algorithmicrequire{\textbf{Input:}}
\algrenewcommand\algorithmicensure{\textbf{Output:}}
\algrenewcommand{\algorithmiccomment}[1]{%
  \hfill{\footnotesize\textcolor{gray}{\textit{#1}}}}
\newcommand{\algheading}[1]{%
  \Statex\vspace{1pt}{\itshape #1}\vspace{1pt}}

\copyrightyear{2027}
\acmYear{2027}
\setcopyright{cc}
\setcctype{by}
\acmConference[WSDM '27]{Proceedings of the Twentieth ACM International Conference on Web Search and Data Mining}{February 15--19, 2027}{Hong Kong, Hong Kong}
\acmBooktitle{Proceedings of the Twentieth ACM International Conference on Web Search and Data Mining (WSDM '27), February 15--19, 2027, Hong Kong, Hong Kong}
\acmDOI{10.1145/3862322.3863725}
\acmISBN{979-8-4007-3117-4/2027/02}

\title{\textsc{FedRCA}: Rank-Conjugate Adaptation for Heterogeneous-Rank Federated LoRA}

\author{Yiming Chen}
\affiliation{%
  \institution{Tsinghua University}
  \city{Beijing}
  \country{China}
}
\email{chenym22@mails.tsinghua.edu.cn}

\author{Zheng Jiang}
\affiliation{%
  \institution{Tsinghua University}
  \city{Beijing}
  \country{China}
}
\email{jz24@mails.tsinghua.edu.cn}

\author{Lifeng Sun}
\authornote{Corresponding author. Lifeng Sun is affiliated with the Department of Computer Science and Technology at Tsinghua University and with the Key Laboratory of Pervasive Computing, Ministry of Education. }
\affiliation{%
  \institution{Tsinghua University}
  \city{Beijing}
  \country{China}
}
\email{sunlf@tsinghua.edu.cn}

\begin{abstract}
Federated low-rank adaptation enables practical fine-tuning of large models without sharing private data or communicating full model updates. In real scenarios, clients may have different LoRA ranks, but existing heterogeneous-rank methods largely treat this difference as a tensor-shape problem. Their rank-specific projections or reconstructions can quietly break a premise of federated averaging: clients should begin each round from the same effective function. Different ranks may instead induce different starting functions and incomparable adaptation coordinates, entangling rank artifacts and leading to uncontrolled parameter updates. We therefore propose FedRCA, a rank-conjugate adaptation protocol that restores a common function-space origin while preserving each client's rank. A fixed spectral atom bank provides deterministic nested coordinates for heterogeneous clients. {For each client, we initialize the rank-specific LoRA adapter from its spectral anchor and subtract the same anchor from the frozen base, so their sum preserves the shared effective model before local training. After training, the server averages the clients' effective-weight displacements from this common origin and broadcasts a compressed low-rank update for merging.} The method requires no additional local training pass, leaves client ranks and uplink shapes unchanged, {and controls the downlink payload through low-rank compression}. Experiments across multiple models and datasets, data partitions, and heterogeneous-rank profiles demonstrate that FedRCA yields consistent gains over strong federated LoRA baselines.
\end{abstract}



\begin{document}
\maketitle

\section{Introduction}

Large language models (LLMs) are increasingly adapted to private data held on user devices and within institutional silos~\cite{zhao2023survey,naveed2025comprehensive,minaee2024large}. Centralizing such data can be impractical or conflict with privacy requirements such as GDPR~\cite{gdpr2018general}; federated learning (FL) instead exchanges model updates while retaining raw records locally~\cite{huang2024federated,wen2023survey,yao2024federated}.

Full LLM training remains expensive, motivating parameter-efficient federated tuning \cite{zhang2024towards,bai2024flexlora}. In particular, LoRA freezes the backbone and trains only low-rank factors $B\in\mathbb{R}^{k_{\mathrm{out}}\times r}$ and $A\in\mathbb{R}^{r\times k_{\mathrm{in}}}$ with effective update $BA$ \cite{hu2022lora}. Existing federated LoRA methods substantially reduce the trainable and communication overhead, but commonly assume that all clients use the same rank~\cite{zhang2024towards,sun2024ffa}. However, clients differ in hardware, latency budgets, and local data complexity, so one LoRA rank may not suit every client. Recent work explores heterogeneous-rank federated LoRA, allowing clients to have different ranks. {Existing heterogeneous-rank methods reconcile adapter shapes through stacking, projection, or rank-specific reconstruction~\cite{bai2024flexlora,wang2024flora}.} This makes aggregation feasible, but does not ensure that the resulting updates are comparable. We treat client ranks as fixed constraints. A client's rank determines how many LoRA parameters it trains and how many factors it sends. The server must reconcile these sizes at aggregation and redistribute a model that each client can adapt at its own rank. Therefore, adapters for clients with lower ranks only receive a subset of the components from the aggregation result, so clients see different effective models at the start of the round. This clearly runs counter to the original intent of FL. Our question is whether redistribution can preserve the shared optimization origin across these adapter sizes.

\begin{figure*}[t]
  \centering
  \includegraphics[width=\textwidth]{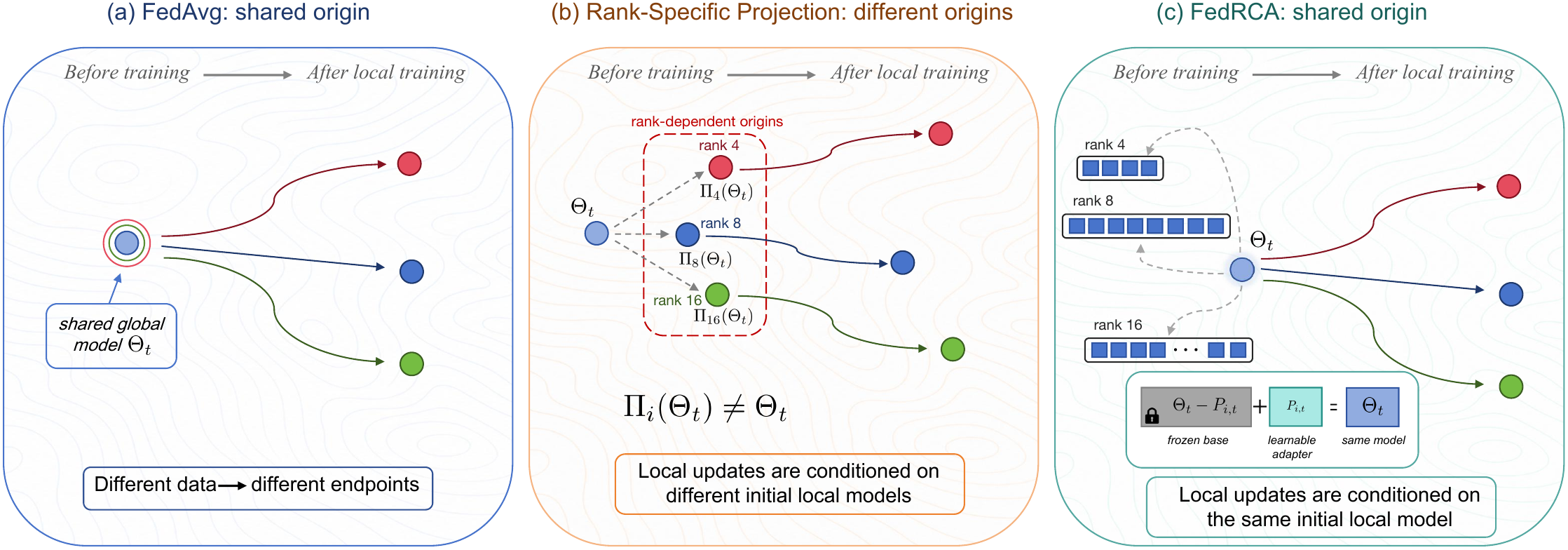}
  \caption{Rank heterogeneity should alter local coordinates, not the initial
  function. \textbf{(a)} FedAvg initializes every client at the shared
  effective model $\Theta_t$; private data may then produce different local
  trajectories. \textbf{(b)} Rank projection initializes client $i$ at
  $\Pi_{r_i}(\Theta_t)\neq\Theta_t$, so its update is conditioned on private
  data and a rank-dependent origin. \textbf{(c)} \method assigns a fixed-prefix anchor $P_{i,t}$ while preserving
  $(\Theta_t-P_{i,t})+P_{i,t}=\Theta_t$. Curves depict conceptual
  optimization trajectories in function space.}
  \Description{Three-panel comparison of standard FedAvg, rank-specific
  projection, and FedRCA. FedAvg and FedRCA begin from one shared
  function, whereas rank-specific projection starts clients from different
  functions before local training.}
  \label{fig:motivation}
\end{figure*}

For projection-based redistribution, the key issue lies in a premise of federated averaging. Standard FedAvg \cite{mcmahan2017fedavg} allows clients to train a model with different private local data, which is local learning, but assumes that all local updates begin from a shared global model. In a projection-based heterogeneous-rank round, however, client $i$ receives a rank-specific reconstruction $\Pi_{r_i}(\Theta_t)$ rather than the shared effective model $\Theta_t$. Let $E_{i,t} = \Pi_{r_i}(\Theta_t) - \Theta_t$ denote the reconstruction error. The client's local optimizer then produces the endpoint
\[
\Theta^{(+)}_{i,t} = \Theta_t + E_{i,t} + \mathcal{U}_i(\Theta_t + E_{i,t};\, \mathcal{D}_i).
\]
Because local optimization starts at $\Theta_t+E_{i,t}$, the projection error $E_{i,t}$ shapes the local training trajectory. Therefore, rank-specific projection can separate their effective models before the first gradient step. As illustrated in Figure~\ref{fig:motivation}(b), rank-specific projection initializes clients at different effective models, so their updates depend on rank-specific starting points in addition to private data and local optimization. We call this function-origin inconsistency. A client that takes no local steps still returns a model with the reconstruction offset $E_{i,t}$. For a client that trains, subtracting this offset from its endpoint leaves the displacement learned along the shifted trajectory. An exact server average therefore combines updates already shaped by different starting models.

This observation suggests a simple principle: rank should determine how a client can learn, not the function from which it starts. We therefore propose \method (\textbf{Fed}erated \textbf{R}ank-\textbf{C}onjugate \textbf{A}daptation). A one-time spectral atom bank gives a rank-$r_i$ client a fixed nested prefix, establishing explicit coordinate correspondence across ranks without a learned router. The corresponding anchor is subtracted from the frozen local base and re-injected through the trainable LoRA factors; the two terms cancel exactly, so every client has the same global function before its first local step. After training, the server removes the anchor, averages only centered function-space displacements, and compresses the result before updating the shared model. 
This construction ensures "no-op consistency": that is, a client that performs no optimization operations has zero impact on the aggregated result.

We evaluate FedRCA on Qwen3-0.6B and Qwen3-4B across arithmetic reasoning, medical question answering, and commonsense reasoning. We cross IID and Dirichlet partitions with two client-rank profiles, both with a mean rank of 16. The strong profile ranges from rank 4 to 64 and the mild profile from rank 8 to 32, varying the distribution of capacity across clients. Extensive experiments have demonstrated that FedRCA achieves the best performance in each combination of model, data partition, and rank profile.

We measure the effective weights before local training to track when clients' starting models diverge. On Qwen3-0.6B under the strong rank profile, FlexLoRA’s models diverge after the first downlink. FedRCA preserves a shared origin throughout. To test its effect on accuracy, we remove anchor cancellation while holding the assigned directions fixed. This yields the largest drop among various factors. In addition, we explored how to construct the initial LoRA factors for each round on the client side. We devise the strategies based on scores and balanced exposure, which will be described in Section~\ref{Mechanism}, and our experiments demonstrate that "fixed spectral ordering" is the optimal approach.

In the compression study, we vary the broadcast rank while keeping each client's adapter rank fixed. A rank-32 broadcast halves the payload of rank 64 and retains over 92\% of the update energy on both models. The mean accuracy changes by about 0.19\% on Qwen3-0.6B and 0.08\% on Qwen3-4B, illustrating the downlink and accuracy trade-off at a fixed client-side budget.

The contributions of this paper are summarized as follows:
\begin{itemize}
    \item We identify function-origin inconsistency as an overlooked failure mode in heterogeneous-rank federated LoRA: rank-specific projection changes both the client's starting function and the local optimization trajectory induced from that point, entangling rank artifacts with genuine local learning.
    \item We propose FedRCA, which combines a fixed spectral atom bank, model-preserving common origins, and rank-conjugate exposure to give arbitrary-rank clients one exact starting function and deterministic nested adaptation subspaces. And we adopt centered function-space aggregation that restores an exact no-op property and preserves the interpretation of federated averaging under heterogeneous LoRA parameterizations.
    \item We conduct extensive experiments across two model scales, multiple tasks, and varying heterogeneity settings, showing that FedRCA achieves consistent improvements over strong baselines. Ablation studies further support the design choices.
\end{itemize}

\section{Related Work}
\label{sec:related}

\subsection{Federated Parameter-Efficient LLM Tuning}
Federated learning (FL) \cite{mcmahan2017fedavg} has been extensively studied across several key dimensions, including generalization \cite{liu2021feddg,chen2024fedawa,chen2025fedtg}, aggregation and optimization strategies \cite{wang2020tackling,karimireddy2020scaffold}, and learning under non-IID data distributions \cite{fedprox}. More recently, the integration of large language models (LLMs) with FL has emerged as an increasingly active research direction. FedIT established federated instruction tuning for LLMs \cite{zhang2024towards}.  FFA-LoRA freezes one factor to remove factor-wise aggregation discordance and reduce uplink size \cite{sun2024ffa}.  FedEx-LoRA adds a residual to make the aggregated effective update exact \cite{singhal2025fedex}.  These methods primarily assume equal factor shapes. Their heterogeneous-rank adaptations are specified in the experimental setup. Communication efficiency also depends on which LoRA factors are shared and when they are updated. FedOTAB alternates which factor is frozen and transmits only the updated factor~\cite{chen2026federated}.

\subsection{Heterogeneous-Rank LoRA}
FlexLoRA averages effective LoRA products, applies an SVD, and redistributes rank-specific prefixes \cite{bai2024flexlora}.  FLoRA stacks heterogeneous factors, exactly represents their weighted effective update, and merges it into the base model \cite{wang2024flora}.  HetLoRA adapts rank and aggregation to on-device model and data heterogeneity \cite{cho2024hetlora}, while ILoRA uses orthogonal initialization and QR aggregation \cite{zhou2025ilora}. Fed-PLoRA reparameterizes LoRA as parallel rank-one modules and uses Select-N-Fold to preserve the full global initialization while different clients train different module subsets \cite{zhang2026fedplora}. PreLort builds a prefix hierarchy with segment-wise aggregation \cite{waseem2026prelort}, and Fed-ALoRA studies asymmetric factor sharing \cite{ban2026fedalora}. \method instead retains conventional rank-\(r_i\) LoRA factors and combines full-function anchor cancellation with a one-time spectral bank, fixed nested exposure, and centered effective-update compression.

\subsection{Model-Heterogeneous Federated Learning}
Broader heterogeneous FL has trained width-varying subnetworks \cite{diao2021heterofl}, exchanged representations through public data and cross-correlation \cite{huang2022fccl}, and aligned global and personalized prompts with heterogeneous prompt lengths \cite{fang2025fedpha}.  \method keeps one frozen LLM backbone, requires no public or probe set, and operates on LoRA-sized client states.  We use randomized SVD \cite{halko2011randomizedsvd} once on the pretrained weights to construct a fixed coordinate bank.

\section{Problem Formulation}
\label{sec:problem}
Consider $K$ clients. Client $i$ owns private data $\mathcal D_i$, receives aggregation weight $p_i=|\mathcal D_i|/\sum_k|\mathcal D_k|$, and trains a rank-$r_i$ LoRA adapter. Let $\mathcal L$ denote the adapted linear layers and $W_t^\ell$ the shared effective weight of layer $\ell$ at the beginning of round $t$.  We collect them as $\Theta_t=\{W_t^\ell\}_{\ell\in\mathcal L}$. Client $i$ uses LoRA scale $s_i=\alpha_i/r_i$ and contribution $s_iB_i^\ell A_i^\ell$, where $\alpha_i$ is its LoRA scaling parameter.

\subsection{Rank-Specific Redistribution Changes the Local Problem}

Let $\Pi_{r_i}$ be a redistribution rule that reconstructs the global adapter at client rank $r_i$. The effective model received by client $i$ is
\begin{equation}
  \Theta_{i,t}^{(0)}
  =\Pi_{r_i}(\Theta_t)
  =\Theta_t+E_{i,t},
  \qquad
  E_{i,t}:=\Pi_{r_i}(\Theta_t)-\Theta_t.
  \label{eq:rank_origin}
\end{equation}
The term $E_{i,t}$ may arise from rank truncation, projection, or a rank-dependent factor prefix. With local optimizer displacement $\mathcal U_i(\Theta;\mathcal D_i)$, the returned model parameter is
\begin{equation}
  \Theta_{i,t}^{(+)}
  =\Theta_t+E_{i,t}
  +\mathcal U_i(\Theta_t+E_{i,t};\mathcal D_i).
  \label{eq:coupled}
\end{equation}
The learned displacement depends on the projected start through its first argument. With the same local optimizer and data, the endpoint from the shared model would be
$\widetilde\Theta_{i,t}^{(+)}
=\Theta_t+\mathcal U_i(\Theta_t;\mathcal D_i)$.  Their difference is
\begin{equation}
  D_{i,t}
  =E_{i,t}
  +\mathcal U_i(\Theta_t+E_{i,t};\mathcal D_i)
  -\mathcal U_i(\Theta_t;\mathcal D_i).
  \label{eq:trajectory_shift}
\end{equation}

If the optimizer map $\mathcal{U}_i(\cdot; \mathcal D_i)$ is locally differentiable around $\Theta_t$, a first-order Taylor expansion of Eq.~\eqref{eq:trajectory_shift} with respect to the initialization gives $D_{i,t} \approx \bigl[I + J_{\mathcal{U}_i}(\Theta_t)\bigr] E_{i,t}.$ This decomposition reveals two distinct sources of deviation. The term $E_{i,t}$ is the direct projection offset: the static discrepancy between the rank-specific reconstruction $\Pi_{r_i}(\Theta_t)$ and the shared model $\Theta_t$ before any local gradient step. The term $J_{\mathcal{U}_i}(\Theta_t)E_{i,t}$ is the trajectory perturbation induced by optimizing from the shifted starting point $\Theta_t + E_{i,t}$ rather than from the true common origin $\Theta_t$. Subtracting $E_{i,t}$ from the trained endpoint can remove only the first, static component. The second component has already been woven into the local trajectory through every gradient computation, parameter update, momentum accumulation, and learning-rate schedule applied to $\Theta_t + E_{i,t}$, so it cannot be eliminated by a post-hoc additive correction. Moreover, the residual $J_{\mathcal{U}_i}(\Theta_t)E_{i,t}$ is intrinsically client-specific, because the Jacobian $J_{\mathcal{U}_i}(\Theta_t)$ depends on the private loss landscape shaped by $\mathcal D_i$, on the feasible parameter subspace determined by rank $r_i$, and on optimizer-specific quantities such as the learning rate, momentum, and number of local steps. Consequently, it cannot be estimated by the server as one correction vector shared by all clients.

Projection-based redistribution can therefore give clients different optimization starting points within the same round. FedAvg allows endpoints to differ with local data, but expects every trajectory to begin at the same global function. Rank-specific redistribution can violate this expectation before any local gradient step.

\subsection{Common-Origin Objective}

We seek a heterogeneous-rank method satisfying
\begin{equation}
  f_{i,t}^{(0)}(x)=f_{\Theta_t}(x),
  \qquad \forall i,\ \forall x,
  \label{eq:common_origin}
\end{equation}
while preserving each client's rank $r_i$. We additionally require no-op consistency: a client that performs no optimization must contribute zero server update. To avoid unrelated rank-specific coordinate systems, we also seek nested exposure sets $\mathcal R_i\subseteq\mathcal R_j$ whenever $r_i\le r_j$. Here \(f\) denotes the deterministic network. These conditions specify the shared origin, the relation between exposed coordinates, and how to measure the trained displacement. Gradients, tangent spaces, and endpoints can vary across clients.

\subsection{From Shape Compatibility to Update Comparability}

Heterogeneous-rank aggregation / reconstruction has three requirements. (1) Shape compatibility requires local LoRA to fit a common server representation. (2) Origin consistency requires updates to be measured from the shared model rather than a rank-specific approximation. (3) Coordinate correspondence requires the directions available at different ranks to be related. Padding, stacking, projection, or factor reconstruction addresses the first requirement, but does not by itself make local updates comparable.

Moreover, LoRA factorization is non-unique. For any invertible matrix \(Q\), \(BA=(BQ)(Q^{-1}A)\), so proximity between factor entries is not a reliable measure of model agreement. We instead define the origin condition in effective-weight and function space. Factor shapes may then differ while the deterministic round-start model remains shared. Rank-conjugate exposure imposes a weaker coordinate condition: shared prefix directions correspond exactly, and higher-rank clients receive additional directions.

\section{The Proposed Method \method}
\label{sec:method}

\begin{figure*}[t]
  \centering
  \includegraphics[width=\textwidth]{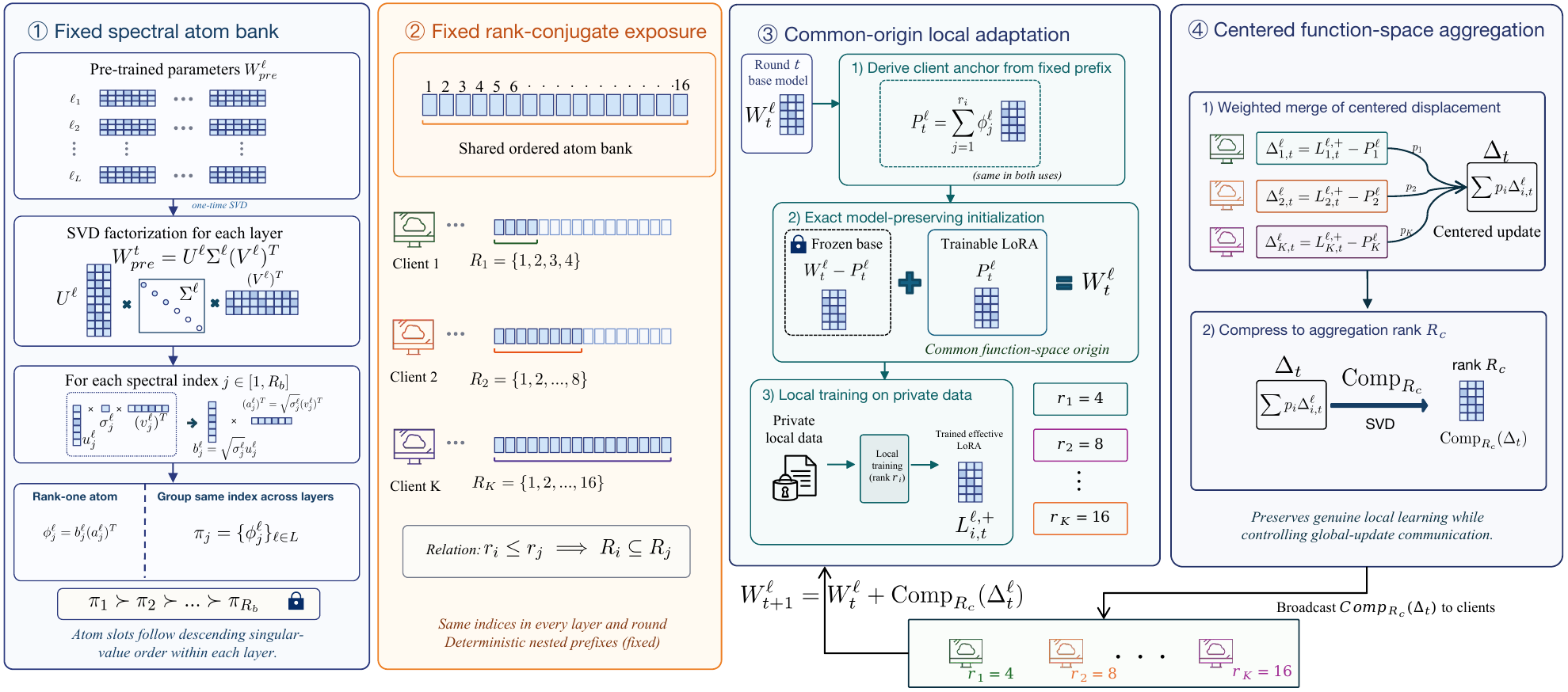}
  \caption{The workflow of \method. \textbf{(1)} A one-time randomized SVD orders each layer's rank-one atoms by singular value; equal spectral indices form model-wide slots. \textbf{(2)} A rank-\(r_i\) client receives the first \(r_i\) slots, producing fixed nested prefixes. \textbf{(3)} Subtracting client anchor \(P_i\) from the frozen base and reintroducing it as the initialized LoRA yields the same deterministic effective weight \(W_t\). \textbf{(4)} The server averages endpoint-minus-anchor displacements, compresses the aggregate to \(R_c\), and broadcasts the factorized update; clients merge it into cached \(W_t\). Together, rank changes learning capacity without changing the deterministic starting function, while compression bounds the shared-update rank.}
  \Description{Four-panel FedRCA workflow. A one-time SVD builds an ordered spectral atom bank; rank-dependent nested prefixes define client anchors; subtracting each anchor from the frozen base and adding it as trainable LoRA gives every client the same deterministic initial model; the server averages centered effective-weight displacements, compresses them, and broadcasts a factorized update.}
  \label{fig:overview}
\end{figure*}

\method gives all clients a common starting function, assigns nested rank-specific adaptation coordinates, and aggregates centered updates. Figure~\ref{fig:overview} summarizes the four-stage protocol. Algorithm~\ref{alg:fedrca} details the training procedure.

\subsection{Fixed Spectral Atom Bank}
Let \(\mathcal L\) denote the LoRA layers. For every pretrained weight \(W_{\mathrm{pre}}^\ell\), the server computes a rank-\(R_b\) randomized SVD once and explicitly sorts its singular values in descending order:
\begin{equation}
  W_{\mathrm{pre}}^\ell
  \approx U^\ell\Sigma^\ell(V^\ell)^\top,\qquad
  \sigma_1^\ell\ge\cdots\ge\sigma_{R_b}^\ell,\qquad
  \phi_j^\ell=b_j^\ell(a_j^\ell)^\top,
  \label{eq:bank}
\end{equation}
where \(b_j^\ell=\sqrt{\sigma_j^\ell}u_j^\ell\) and \((a_j^\ell)^\top=\sqrt{\sigma_j^\ell}(v_j^\ell)^\top\). Hence \(\|\phi_j^\ell\|_F=\sigma_j^\ell\): within each projection, atoms are ordered by descending rank-one Frobenius norm, equivalently singular value. We group the same spectral index across all adapted projection layers as
\begin{equation}
  \pi_j:=\{\phi_j^\ell\}_{\ell\in\mathcal L},
  \qquad
  \mathcal A=\{\pi_1,\ldots,\pi_{R_b}\}.
  \label{eq:model_atom}
\end{equation}
Each \(\pi_j\) packs the $j$-th component from every layer. Components are not globally reordered across layers. We retain the fixed index order \((\pi_1,\ldots,\pi_{R_b})\), with \(R_b\ge\max_i r_i\), before FL training and independent of client data, local updates and other factors.

\subsection{Fixed Rank-Conjugate Exposure}
A client of rank \(r_i\) receives the deterministic prefix
\begin{equation}
  \mathcal R_i=\{1,\ldots,r_i\}.
  \label{eq:prefix_route}
\end{equation}
The same indices are used at every round and in every LoRA layer. For any \(r_i\le r_j\),
\begin{equation}
  \mathcal R_i\subseteq\mathcal R_j.
  \label{eq:nested_exposure}
\end{equation}
Therefore, spectral position \(k\) is available to exactly those clients with \(r_i\ge k\). We call this nesting rank-conjugate exposure. Lower-rank clients optimize the most widely shared prefix, while higher-rank clients retain that prefix and extend it with additional directions.

Different ranks still expose different numbers of trainable directions. The nesting only prevents them from using unrelated coordinate systems: every lower-rank prefix is contained in each compatible higher-rank prefix. The fixed assignment costs \(O(\sum_i r_i)\) to construct and has no routing hyperparameter. At the same time, this approach ensures that the directions of LoRA with different ranks remain as consistent as possible.

\subsection{Model-Preserving Common Function-Space Origin}

For client \(i\), the effective anchor in projection \(\ell\) is
\begin{equation}
  P_i^\ell=\sum_{j=1}^{r_i}\phi_j^\ell.
  \label{eq:fixed_anchor}
\end{equation}
For a client whose rank \(r_i\) stays fixed, the anchor \(P_i^\ell\) is the same in every communication round. Initialize its rank-\(r_i\) LoRA factors as
\begin{align}
  B_{i,t}^{\ell,(0)}
  &=[b_1^\ell,\ldots,b_{r_i}^\ell],\label{eq:factors_B}\\
  A_{i,t}^{\ell,(0)}
  &=\frac{1}{s_i}
    [(a_1^\ell)^\top;\ldots;(a_{r_i}^\ell)^\top],
  \label{eq:factors}
\end{align}
where \(s_i=\alpha_i/r_i\) is the client LoRA scale. Then
\(s_iB_{i,t}^{\ell,(0)}A_{i,t}^{\ell,(0)}=P_i^\ell\).

Using \(P_i^\ell\) directly as a rank-specific approximation would bring the origin mismatch again. \method instead subtracts the anchor from the frozen base and adds it back through the trainable adapter:
\begin{equation}
  \underbrace{(W_t^\ell-P_i^\ell)}_{\text{frozen local base}}
  +\underbrace{s_iB_{i,t}^{\ell,(0)}
  A_{i,t}^{\ell,(0)}}_{\text{trainable LoRA}}
  =W_t^\ell.
  \label{eq:cancellation}
\end{equation}
Factor shapes and exposed coordinates may differ, but every client's effective model is exactly \(\Theta_t\), thereby resolving the issue of inconsistent origin functions mentioned earlier.

\paragraph{Proposition 1 (common origin and exact no-op).} For every rank and round, Eq.~\eqref{eq:cancellation} gives \(\Theta_{i,t}^{(0)}\equiv\Theta_t\). If a client performs no local parameter update, its centered contribution to the server is exactly zero.

\textit{Proof.} Equations~\eqref{eq:factors_B}--\eqref{eq:factors} make the initialized LoRA product equal \(P_i^\ell\), which cancels layer by layer in Eq.~\eqref{eq:cancellation}. If the factors do not change, the post-training LoRA product remains \(P_i^\ell\); subtracting the anchor therefore returns zero. \hfill\(\square\)

\subsection{Centered Effective-Update Aggregation}

After local training, the local LoRA is
\begin{equation}
  L_{i,t}^{\ell,(+)}
  =s_iB_{i,t}^{\ell,(+)}A_{i,t}^{\ell,(+)}.
\end{equation}
The displacement from the initialized LoRA is
\begin{equation}
  \Delta_{i,t}^\ell
  :=L_{i,t}^{\ell,(+)}-P_i^\ell.
  \label{eq:centered}
\end{equation}
Therefore, the client's effective endpoint is
\begin{equation}
  (W_t^\ell-P_i^\ell)+L_{i,t}^{\ell,(+)}
  =W_t^\ell+\Delta_{i,t}^\ell.
  \label{eq:endpoint}
\end{equation}
Unlike a post-hoc subtraction of the projection error in Eq.~\eqref{eq:coupled}, this displacement was actually trained from the common function \(\Theta_t\). It still reflects private data, rank, and reachable subspace, but not a rank-specific initial function.

The server aggregates
\begin{align}
  {\Delta}_t^\ell
    &=\sum_{i=1}^{K}p_i\Delta_{i,t}^\ell,\\
  W_{t+1}^\ell
    &=W_t^\ell+\mathsf{Comp}_{R_c}
      ({\Delta}_t^\ell).
  \label{eq:aggregation}
\end{align}
The sum of sample weights is 1, so averaging the full client endpoints in Eq.~\eqref{eq:endpoint} would recover \(W_t^\ell+\Delta_t^\ell\) before compression. The centered representation lets the server compute the same weighted aggregate directly from the adapter factors and their anchors.

The \textit{aggregation rank} \(R_c\) specifies how much of the aggregated displacement is retained. The atom-bank rank \(R_b\) determines the spectral anchors available at initialization. In the main experiments we set \(R_c=R_b\). To compress the aggregation result, the server concatenates pairs of skinny factors for the trained adapters and anchors. The aggregation weights and the negative signs for the anchors can be absorbed into one side of each factor product. QR factorizations then reduce the SVD to a small core, whose leading components form the compressed update. Compression follows client weighting; the server then broadcasts the factorized update \(\widehat\Delta_t=\mathsf{Comp}_{R_c}(\Delta_t)\), which clients merge into base model \(W_t\). Each layer holds the shared effective weight after merging the broadcast update. At the next round, each client uses its fixed anchor to split this weight into a frozen base and an initialized adapter. The two parts again sum to the same shared model for every client, while the higher-rank adapters expose additional directions.

\begin{algorithm}[t]
\caption{Training procedure of \method}
\label{alg:fedrca}
\begin{algorithmic}[1]
\Require Models \(\Theta_{\mathrm{pre}},\Theta_0\); clients
\(\{(\mathcal D_i,p_i,r_i)\}_{i=1}^{K}\); rounds \(T\); local epochs \(E\);
ranks \(R_b,R_c\)
\Ensure Shared effective model \(\Theta_T\)

\algheading{One-time setup}
\State \(\mathcal A\gets
  \operatorname{SpectralBank}_{R_b}(\Theta_{\mathrm{pre}})\)
  \Comment{rSVD, computed once}
\ForAll{clients \(i=1,\ldots,K\)}
  \State \(\mathcal R_i\gets\{1,\ldots,r_i\}\)
  \Comment{nested prefix}
\EndFor
\State Cache \(\Theta_0\), \(\mathcal A\), and
\(\{\mathcal R_i\}_{i=1}^{K}\)

\algheading{Federated rounds}
\For{\(t=0,\ldots,T-1\)}
  \ForAll{clients \(i\) \textbf{in parallel}}
    \ForAll{adapted layers \(\ell\)}
      \State \(P_i^\ell\gets\sum_{j\in\mathcal R_i}\phi_j^\ell\), \quad
      \(W_{i,t}^{\ell,\mathrm{base}}\gets W_t^\ell-P_i^\ell\)
      \State Choose \((B_{i,t}^{\ell,(0)},A_{i,t}^{\ell,(0)})\) such that
      \(s_iB_{i,t}^{\ell,(0)}A_{i,t}^{\ell,(0)}=P_i^\ell\)
      \State \(W_{i,t}^{\ell,\mathrm{base}}+
        s_iB_{i,t}^{\ell,(0)}A_{i,t}^{\ell,(0)}=W_t^\ell\)
        \Comment{deterministic common origin}
    \EndFor
    \State \((B_{i,t}^{(+)},A_{i,t}^{(+)})\gets
      \operatorname{LocalTrain}_{E}(\mathcal D_i;
      W_{i,t}^{\mathrm{base}},B_{i,t}^{(0)},A_{i,t}^{(0)})\)
    \State Upload endpoint factors
      \((B_{i,t}^{(+)},A_{i,t}^{(+)})\)
  \EndFor

  \ForAll{adapted layers \(\ell\)} \Comment{on server}
    \State \(\Delta_{i,t}^\ell\gets L_{i,t}^{\ell,(+)}-P_i^\ell\),
      \quad \(\forall i\)
      \Comment{remove anchor}
    \State \({\Delta}_t^\ell\gets\sum_i p_i\Delta_{i,t}^\ell\)
    \State \(\widehat\Delta_t^\ell\gets
      \mathsf{Comp}_{R_c}({\Delta}_t^\ell)\)
      \Comment{skinny QR/core-SVD}
    \State \(W_{t+1}^\ell\gets W_t^\ell+\widehat\Delta_t^\ell\)
  \EndFor
  \State Broadcast factorized \(\widehat\Delta_t\); clients merge it into
    cached \(\Theta_t\)
\EndFor
\end{algorithmic}
\end{algorithm}

\subsection{Complexity and Communication Scope}

Relative to ordinary rank-\(r_i\) federated LoRA, \method does not change the client rank, local forward/backward computation, number of local epochs, or adapter uplink shape. It requires no additional client-side data pass. The fixed routes contain \(\sum_i r_i\) integer indices and can be constructed once in \(O(\sum_i r_i)\) time. 

The uplink is unchanged, but the downlink is larger. Let
\begin{equation}
  P(r)=r\sum_{\ell\in\mathcal L}
  (d_{\mathrm{in}}^\ell+d_{\mathrm{out}}^\ell)
  \label{eq:lora_size}
\end{equation}
be the number of scalars in a rank-\(r\) LoRA over all adapted layers. A rank-specific redistribution baseline uploads and downloads \(\sum_iP(r_i)\) scalars per round. \method has the same uplink, but every client receives the common factorized rank-\(R_c\) update, giving client-wise downlink traffic
\begin{equation}
  C_{\mathrm{down}}^{\method}=K P(R_c),\qquad
  \frac{C_{\mathrm{down}}^{\method}}
  {C_{\mathrm{down}}^{\mathrm{rank\text{-}specific}}}
  =\frac{R_c}{\bar r},
  \label{eq:downlink_ratio}
\end{equation}
where \(\bar r=K^{-1}\sum_i r_i\). Table~\ref{tab:communication} instantiates this accounting for the main profiles. The \(r_i\) integer route indices are fixed and cached; the full bank can be reconstructed from the pretrained weights or transmitted once. Neither is a per-round payload.

\begin{table}[t]
\centering
\caption{Per-round aggregate communication (MiB) with FP16 \(q/v\) LoRA and
client-wise unicast accounting. ``Base DL'' sends rank-\(r_i\) LoRA; \method
sends the rank-\(R_c\) LoRA to every client.}
\label{tab:communication}
\scriptsize
\setlength{\tabcolsep}{2.1pt}
\begin{tabular}{llrrrr}
\toprule
Model & Profile & Uplink & Base DL & \method DL & Ratio\\
\midrule
0.6B & Strong, \(R_c=64\) & 43.75 & 43.75 & 175.00 & 4.0\(\times\)\\
0.6B & Mild, \(R_c=32\) & 43.75 & 43.75 & 87.50 & 2.0\(\times\)\\
4B & Strong, \(R_c=64\) & 112.50 & 112.50 & 450.00 & 4.0\(\times\)\\
4B & Mild, \(R_c=32\) & 112.50 & 112.50 & 225.00 & 2.0\(\times\)\\
\bottomrule
\end{tabular}
\end{table}

\method trades additional downlink traffic for a common starting model. The numbers count a common broadcast once per receiving client, as is conventional in FL communication analyses.

\section{Experiments}
\label{sec:experiments}

Our evaluation measures effective-model origin discrepancy, compares task accuracy with baselines across models and heterogeneity settings, {assesses the method's design choices}, and examines the relation between aggregation rank, retained update energy, and accuracy.

\subsection{Experimental Setup}
\paragraph{Models and datasets.}
The main study fine-tunes Qwen3-0.6B and Qwen3-4B \cite{yang2025qwen3} on GSM8K \cite{cobbe2021gsm8k}, MedQA-USMLE \cite{jin2021medqa}, and CommonsenseQA \cite{talmor2019commonsenseqa}. The tasks cover free-form arithmetic, medical multiple-choice knowledge, and commonsense multiple-choice reasoning. Table~\ref{tab:datasets} summarizes the splits.

\begin{table}[t]
\centering
\caption{Datasets used in the main comparison. Training examples are federated across clients; evaluation uses the complete held-out split.}
\label{tab:datasets}
\small
\setlength{\tabcolsep}{4.2pt}
\begin{tabular}{lrrl}
\toprule
Dataset & Train & Eval & Partition label\\
\midrule
GSM8K & 7,473 & 1,319 & difficulty quintile\\
MedQA & 10,178 & 1,273 & answer option\\
CommonsenseQA & 9,741 & 1,221 & answer option\\
\bottomrule
\end{tabular}
\end{table}

\paragraph{Federated partitions.}
Every run uses \(K=10\) fully participating clients, and all compared methods reuse the same serialized partition for a given dataset and seed. IID partitions are balanced by sample count. Non-IID partitions use a label-aware Dirichlet allocation with \(\alpha=0.1\) and at least 32 examples per client. MedQA and CommonsenseQA use the gold answer option as the partition label. Because numeric GSM8K answers are mostly singletons, GSM8K uses equal-frequency reasoning-difficulty quintiles obtained by lexicographically sorting examples by annotated solution-step count, annotated-solution length, and question length. Seeds \(\{1,2,3\}\) control partitioning, seeded randomized-SVD bank construction (where applicable), and optimization.

\paragraph{Controlled rank profiles.}
Let \(r^{\times m}\) denote \(m\) clients of rank \(r\). The strong profile is \([4^{\times4},8^{\times2},16^{\times2},32,64]\), with \(R_b=R_c=64\). The mild profile is \([8^{\times4},16^{\times4},32^{\times2}]\), with \(R_b=R_c=32\). Both profiles have mean client rank 16. They change rank dispersion together with the matched atom-bank and aggregation ranks while keeping total client-side LoRA rank fixed. By Eq.~\eqref{eq:lora_size}, the total LoRA parameter count for a given model is proportional to the sum of client ranks. The two profiles therefore match aggregate uplink and assign different adapter budgets to individual clients.

\paragraph{Optimization.}
All methods use the same data order and AdamW optimizer with 3\% linear warmup and zero weight decay within a matched run. Clients train three local epochs with learning rate \(2\times10^{-4}\), effective batch size 32, LoRA dropout 0.05, scale \(s_i=2\), maximum sequence length 512, and adapters on the attention \(q\) and \(v\) projections. Qwen3-0.6B uses micro-batch size 4 with eight accumulation steps; Qwen3-4B uses micro-batch size 2 with 16 steps to preserve the same effective batch. Evaluation batch size is 16 and training uses FP16. We use Alpaca-style prompts with response-only loss and greedy decoding (128 new tokens for GSM8K and 8 for multiple-choice tasks). We evaluate every method at round 10 (R10) for Qwen3-0.6B and at round 5 (R5) for Qwen3-4B. We report mean \(\pm\) sample standard deviation over three seeds.

\paragraph{Baselines.}
We compare FedIT \cite{zhang2024towards}, FFA-LoRA \cite{sun2024ffa}, FlexLoRA \cite{bai2024flexlora}, FedEx-LoRA \cite{singhal2025fedex}, FLoRA \cite{wang2024flora}, and Fed-PLoRA \cite{zhang2026fedplora}. FlexLoRA, FLoRA, and Fed-PLoRA natively accept heterogeneous ranks. FedIT, FFA-LoRA, and FedEx-LoRA assume equal-shaped factors in their original protocols. For these methods, we embed client factors into server-rank nested prefixes, apply the original aggregation rule, and redistribute client-rank prefixes. Every method receives the same ranks, partitions, local optimization, precision, target modules, seeds, and reporting round.

Our PEFT reproduction of Fed-PLoRA fuses selected one-rank paths into one rank-\(r_i\) adapter and uses the shared scale/dropout setup; it is therefore an implementation-aligned control rather than a paper-native reproduction. Its contributor-only component mean, \(\lvert Q_j\rvert^{-1}\sum_{i\in Q_j}\Delta\theta_{i,j}\), where \(Q_j=\{i:j\in\mathcal R_i\}\), collapsed in the strong profile, coinciding with large updates on sparsely exposed components. Replacing \(\lvert Q_j\rvert\) by \(K\) stabilized this implementation: we treat non-contributors as zero and report \(K^{-1}\sum_{i\in Q_j}\Delta\theta_{i,j}\), marked \(\ddagger\) and shown in gray.

\begin{figure}[t]
  \centering
  \includegraphics[width=\columnwidth]{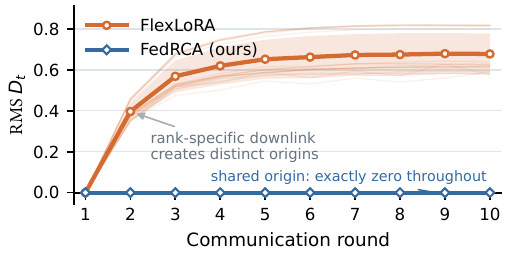}
  \caption{Start-of-round effective-weight origin discrepancy (Eq.~\eqref{eq:origin_discrepancy}) on an 18-run strong-profile ensemble (ten-client Qwen3-0.6B; mean \(\pm\) standard deviation; faint paths show individual runs). FlexLoRA diverges after rank-specific downlink; \method remains at zero by construction.}
  \Description{A line plot over ten communication rounds. FlexLoRA starts at
  zero discrepancy, rises sharply after the first aggregation, and remains
  near 0.68 in later rounds. FedRCA remains exactly at zero in every round.}
  \label{fig:origin-discrepancy}
\end{figure}

\paragraph{Origin-discrepancy diagnostic.}
Figure~\ref{fig:origin-discrepancy} reports the sample-weighted pairwise RMS Frobenius distance between clients' adapted deterministic start-of-round effective weights:
\begin{equation}
  D_t=\left[\sum_{i,j}p_ip_j
  \left\|\Theta_{i,t}^{(0)}-\Theta_{j,t}^{(0)}\right\|_F^2\right]^{1/2}.
  \label{eq:origin_discrepancy}
\end{equation}
Equal starting weights give the same deterministic function. A nonzero \(D_t\) measures a discrepancy in the clients’ starting weights. We reconstruct FlexLoRA origins by replaying previous-round endpoints through its unchanged aggregation and downlink; its shared base cancels pairwise, reducing \(D_t\) to the adapter-operator distance. For \method, the base and anchor cancel, so \(D_t=0\) algebraically. Across 18 runs, FlexLoRA rises from zero at R1 to 0.396 at R2, 0.652 at R5, and 0.678 at R10, while all 180 \method round records remain at zero.

\subsection{Main Results}
\label{sec:main-results}

\begin{table*}[t]
\centering
\caption{Qwen3-0.6B accuracy (\%) at fixed R10. S/M denote strong/mild heterogeneous-rank profiles. Values are mean \(\pm\) sample standard deviation over three seeds; Avg is the unweighted mean across the three tasks. \(\ddagger\) marks the implementation-aligned Fed-PLoRA control, whose values are shown in gray and excluded from ranking. Bold and underlined values denote the best and second-best mean accuracies in each column and rank profile.}
\label{tab:main06b}
\footnotesize
\setlength{\tabcolsep}{2.5pt}
\renewcommand{\arraystretch}{1.06}
\begin{tabular}{llcccccccc}
\toprule
& & \multicolumn{4}{c}{IID} & \multicolumn{4}{c}{Dirichlet \(\alpha=0.1\)}\\
\cmidrule(lr){3-6}\cmidrule(lr){7-10}
Rank & Method & GSM8K & MedQA & CSQA & Avg & GSM8K & MedQA & CSQA & Avg\\
\midrule
\multirow{7}{*}{S}
& FedIT & \acc{35.91}{0.61} & \acc{34.83}{0.46} & \acc{64.51}{0.70} & 45.08
& \acc{36.62}{0.66} & \acc{30.71}{2.79} & \acc{61.37}{0.42} & 42.90\\
& FFA-LoRA & \acc{35.08}{0.95} & \second{\acc{36.03}{0.68}} & \acc{63.06}{0.85} & 44.72
& \acc{35.13}{1.17} & \acc{31.19}{3.16} & \acc{56.54}{2.37} & 40.95\\
& FlexLoRA & \second{\acc{37.20}{0.16}} & \acc{34.46}{0.60} & \second{\acc{64.81}{0.25}} & \second{45.49}
& \second{\acc{37.68}{0.85}} & \best{\acc{32.34}{0.75}} & \second{\acc{63.80}{0.81}} & \second{44.61}\\
& FedEx-LoRA & \acc{36.95}{0.80} & \acc{33.88}{0.82} & \acc{63.66}{0.26} & 44.83
& \acc{37.50}{0.52} & \acc{29.41}{2.43} & \acc{60.61}{0.59} & 42.51\\
& FLoRA & \acc{36.64}{0.57} & \acc{34.15}{0.71} & \acc{63.36}{0.09} & 44.72
& \acc{36.95}{0.34} & \acc{30.16}{3.19} & \acc{60.58}{0.48} & 42.56\\
& Fed-PLoRA\(^{\ddagger}\) & \stabilized{\acc{37.17}{1.08}} & \stabilized{\acc{34.51}{0.20}} & \stabilized{\acc{63.72}{0.43}} & \stabilized{45.13}
& \stabilized{\acc{37.12}{0.49}} & \stabilized{\acc{31.21}{2.27}} & \stabilized{\acc{60.44}{0.89}} & \stabilized{42.93}\\
\rowcolor{mosaiclight}
& \method & \best{\acc{38.39}{0.68}} & \best{\acc{37.78}{1.09}} & \best{\acc{65.49}{1.18}} & \best{47.22}
& \best{\acc{38.41}{0.74}} & \second{\acc{32.08}{3.63}} & \best{\acc{64.32}{0.13}} & \best{44.94}\\
\midrule
\multirow{7}{*}{M}
& FedIT & \acc{36.42}{0.61} & \second{\acc{36.21}{1.39}} & \second{\acc{64.70}{1.21}} & 45.78
& \acc{37.43}{0.44} & \second{\acc{32.39}{2.17}} & \acc{61.64}{0.76} & 43.82\\
& FFA-LoRA & \acc{35.71}{0.66} & \acc{34.09}{0.89} & \acc{63.20}{0.79} & 44.33
& \acc{36.62}{0.66} & \acc{28.99}{1.86} & \acc{58.67}{1.11} & 41.42\\
& FlexLoRA & \second{\acc{37.28}{0.81}} & \acc{35.64}{1.18} & \acc{64.54}{0.75} & \second{45.82}
& \second{\acc{37.88}{1.03}} & \acc{32.16}{1.19} & \second{\acc{62.33}{0.59}} & \second{44.12}\\
& FedEx-LoRA & \acc{36.26}{0.42} & \acc{34.77}{0.05} & \acc{63.77}{0.21} & 44.94
& \acc{37.17}{0.97} & \acc{31.84}{0.40} & \acc{60.96}{1.84} & 43.33\\
& FLoRA & \acc{37.10}{0.74} & \acc{32.84}{1.26} & \acc{62.08}{0.36} & 44.01
& \acc{37.71}{0.50} & \acc{30.85}{0.28} & \acc{59.73}{1.37} & 42.76\\
& Fed-PLoRA\(^{\ddagger}\) & \stabilized{\acc{36.82}{0.69}} & \stabilized{\acc{33.99}{0.40}} & \stabilized{\acc{63.75}{0.62}} & \stabilized{44.85}
& \stabilized{\acc{37.38}{0.27}} & \stabilized{\acc{31.71}{0.24}} & \stabilized{\acc{61.81}{1.32}} & \stabilized{43.63}\\
\rowcolor{mosaiclight}
& \method & \best{\acc{37.91}{0.66}} & \best{\acc{38.31}{0.56}} & \best{\acc{65.44}{0.00}} & \best{47.22}
& \best{\acc{38.99}{1.03}} & \best{\acc{34.28}{2.73}} & \best{\acc{63.88}{0.38}} & \best{45.72}\\
\bottomrule
\end{tabular}
\end{table*}

Tables~\ref{tab:main06b} and~\ref{tab:main4b} report accuracy across tasks, partitions, and rank profiles. Among ranked methods, \method leads 21 of the 24 comparisons and attains the best average in all eight partition and profile blocks. Its overall accuracy is \(46.27\%\) versus \(45.01\%\) for FlexLoRA in the 0.6B/R10 evaluation, and \(66.60\%\) versus \(65.48\%\) for FedIT in the 4B/R5 evaluation. These are absolute gains of {\(1.27\%\)} and {\(1.12\%\)} over the strongest baseline. Relative to the native heterogeneous-rank baseline FlexLoRA, the corresponding seed-matched outcomes are 29/2/5 and 33/0/3 (wins/ties/losses), respectively, totaling 62 wins, two ties, and eight losses over 72 pairs. {The gains span tasks, partitions, and rank profiles.} {Reported accuracy changes are absolute differences.} These differences use the unrounded means underlying the tables. The three non-leading cells are localized. In strong-profile Dirichlet MedQA at 0.6B/R10, FlexLoRA reaches \(32.34\%\pm0.75\%\), compared with \(32.08\%\pm3.63\%\) for \method. At 4B/R5, \method trails {FedIT by} {\(0.07\%\)} on IID-strong GSM8K and by {\(0.21\%\)} on Dirichlet-mild MedQA.

\begin{table*}[t]
\centering
\caption{Qwen3-4B accuracy (\%) at fixed R5. S/M denote strong/mild heterogeneous-rank profiles. Values are mean \(\pm\) sample standard deviation
over three seeds; Avg is the unweighted mean across the three tasks.
\(\ddagger\) marks the implementation-aligned Fed-PLoRA control, whose values are
shown in gray and excluded from ranking. Bold and underlined values denote
the best and second-best mean accuracies in each column and rank profile.}
\label{tab:main4b}
\footnotesize
\setlength{\tabcolsep}{2.5pt}
\renewcommand{\arraystretch}{1.06}
\begin{tabular}{llcccccccc}
\toprule
& & \multicolumn{4}{c}{IID} & \multicolumn{4}{c}{Dirichlet \(\alpha=0.1\)}\\
\cmidrule(lr){3-6}\cmidrule(lr){7-10}
Rank & Method & GSM8K & MedQA & CSQA & Avg & GSM8K & MedQA & CSQA & Avg\\
\midrule
\multirow{7}{*}{S}
& FedIT & \best{\acc{56.58}{0.44}} & \second{\acc{59.62}{0.24}} & \second{\acc{82.15}{0.62}} & \second{66.12}
& \acc{57.04}{0.44} & \second{\acc{55.56}{4.26}} & \acc{80.67}{0.30} & 64.43\\
& FFA-LoRA & \acc{55.83}{0.49} & \acc{58.86}{0.12} & \acc{79.61}{0.30} & 64.77
& \acc{56.33}{0.93} & \acc{51.14}{5.72} & \acc{76.82}{0.71} & 61.43\\
& FlexLoRA & \acc{54.94}{0.39} & \acc{59.33}{0.05} & \acc{81.82}{0.67} & 65.37
& \acc{56.89}{0.96} & \second{\acc{55.56}{3.27}} & \second{\acc{81.24}{0.36}} & \second{64.57}\\
& FedEx-LoRA & \acc{55.85}{1.03} & \acc{58.50}{0.61} & \acc{81.57}{0.71} & 65.31
& \acc{56.58}{0.69} & \acc{49.44}{9.35} & \acc{80.67}{0.46} & 62.23\\
& FLoRA & \acc{56.43}{1.03} & \acc{59.41}{0.25} & \acc{80.89}{0.34} & 65.58
& \second{\acc{57.19}{0.31}} & \acc{52.95}{7.33} & \acc{80.29}{0.98} & 63.48\\
& Fed-PLoRA\(^{\ddagger}\) & \stabilized{\acc{56.61}{0.46}} & \stabilized{\acc{60.04}{0.53}} & \stabilized{\acc{81.65}{0.33}} & \stabilized{66.10}
& \stabilized{\acc{57.14}{0.83}} & \stabilized{\acc{52.74}{8.57}} & \stabilized{\acc{79.55}{0.85}} & \stabilized{63.14}\\
\rowcolor{mosaiclight}
& \method & \second{\acc{56.51}{0.69}} & \best{\acc{61.17}{0.35}} & \best{\acc{82.64}{0.43}} & \best{66.77}
& \best{\acc{57.70}{0.92}} & \best{\acc{59.75}{1.14}} & \best{\acc{82.09}{0.66}} & \best{66.51}\\
\midrule
\multirow{7}{*}{M}
& FedIT & \second{\acc{56.43}{0.68}} & \acc{59.02}{0.63} & \second{\acc{81.85}{0.13}} & \second{65.77}
& \second{\acc{56.94}{0.55}} & \best{\acc{58.44}{2.72}} & \second{\acc{81.46}{0.17}} & \second{65.62}\\
& FFA-LoRA & \acc{55.83}{0.77} & \second{\acc{60.36}{0.32}} & \acc{80.26}{0.25} & 65.48
& \acc{56.51}{0.70} & \acc{55.51}{2.13} & \acc{76.99}{1.01} & 63.00\\
& FlexLoRA & \acc{54.94}{0.46} & \acc{59.05}{0.60} & \acc{81.63}{0.09} & 65.21
& \acc{56.33}{0.40} & \acc{56.79}{2.38} & \acc{81.19}{0.50} & 64.77\\
& FedEx-LoRA & \acc{55.77}{0.19} & \acc{58.00}{0.65} & \acc{81.33}{0.28} & 65.03
& \acc{56.33}{0.66} & \acc{56.87}{2.27} & \acc{81.33}{0.54} & 64.84\\
& FLoRA & \acc{55.72}{0.13} & \acc{59.10}{0.43} & \acc{80.78}{0.25} & 65.20
& \acc{56.86}{0.55} & \acc{51.69}{8.44} & \acc{79.99}{1.31} & 62.85\\
& Fed-PLoRA\(^{\ddagger}\) & \stabilized{\acc{56.13}{0.31}} & \stabilized{\acc{59.26}{0.51}} & \stabilized{\acc{81.63}{0.59}} & \stabilized{65.67}
& \stabilized{\acc{56.25}{0.27}} & \stabilized{\acc{58.16}{1.40}} & \stabilized{\acc{80.97}{0.25}} & \stabilized{65.13}\\
\rowcolor{mosaiclight}
& \method & \best{\acc{56.76}{0.68}} & \best{\acc{61.51}{0.67}} & \best{\acc{83.21}{0.33}} & \best{67.16}
& \best{\acc{57.62}{0.33}} & \second{\acc{58.24}{2.60}} & \best{\acc{82.01}{0.47}} & \best{65.96}\\
\bottomrule
\end{tabular}
\end{table*}

\subsection{Robustness Across Tasks and Heterogeneity}

\paragraph{Where the gains arise.}
To separate the overall performance from task-specific behavior, we compare \method with FlexLoRA, the strongest native heterogeneous-rank baseline, after averaging each task over the four combinations of partition and rank profile. On Qwen3-0.6B, the absolute gains are {\(0.92\%\)} on GSM8K, {\(1.96\%\)} on MedQA, and {\(0.91\%\)} on CommonsenseQA. On Qwen3-4B, they are {\(1.37\%\)},
{\(2.49\%\)}, and {\(1.02\%\)}, respectively. The improvement is present on all three tasks in both
evaluations, but is consistently largest on MedQA. The larger MedQA gains suggest that knowledge-intensive tasks may be
particularly sensitive to misaligned starting models.

\paragraph{Data heterogeneity and rank dispersion.}
Averaged over the two rank profiles, the accuracy of \method drops from \(47.22\%\)
under IID to \(45.33\%\) under Dirichlet partitioning on Qwen3-0.6B, an
absolute drop of {\(1.89\%\)}. On Qwen3-4B, the corresponding values are
\(66.97\%\) and \(66.24\%\), an absolute
drop of {\(0.73\%\)} in the
4B/R5 setting. \method achieves the highest average accuracy in all four
combinations of partition and rank profile at both scales. The strong and mild profiles,
which jointly vary rank dispersion and the matched \(R_b/R_c\), differ by only
{\(0.39\%\)} on Qwen3-0.6B and {\(0.08\%\)} on Qwen3-4B. For FlexLoRA, the corresponding
differences are {\(0.08\%\)} and {\(0.02\%\)}. Both profiles have the same mean adapter rank, so the gains occur at matched
mean adapter capacity.

\paragraph{Non-IID stability.}
The differences in performance among the various methods are most pronounced on the MedQA dataset under the "strong profile" and Dirichlet data partitioning settings. On Qwen3-0.6B, FlexLoRA leads \method by \(0.26\%\) and has a lower standard deviation across seeds. On Qwen3-4B, however, \method reaches \(59.75\%\pm1.14\%\), compared with \(55.56\%\pm3.27\%\) for FlexLoRA; the other baselines have standard deviations between 4.26 and 9.35 in the same cell. At 4B/R5, the improvement includes both a higher mean and a lower observed seed standard deviation under this difficult split. On mild-profile Dirichlet MedQA at 4B/R5, FedIT leads \method by \(0.21\%\).

\subsection{Mechanism Analysis}
\label{Mechanism}
Table~\ref{tab:mechanism_evidence} compares \method (Fixed Order) with three variants. Score Order reorders fixed atoms using round-wise consensus relevance. Balanced Exposure uses the same relevance estimates while balancing atom exposure across clients. No Common Origin replays the same Score-Order exposure sets but removes anchor cancellation from the frozen base{, leaving the spectral anchor uncanceled at initialization}.

\begin{table}[t]
\centering
\caption{Accuracy (\%) at the shared final checkpoint, averaged
over tasks, partitions, rank profiles, and seeds. Balanced Exposure and No Common Origin are matched controls for Score Order.}
\label{tab:mechanism_evidence}
\scriptsize
\setlength{\tabcolsep}{2.0pt}
\begin{tabular}{llllrr}
\toprule
Variant & Order & Exposure & Origin & 0.6B & 4B\\
\midrule
\rowcolor{mosaiclight}
Fixed Order & Fixed & Rank-conj. & Yes & \best{46.273} & \best{66.600}\\
Score Order & Score & Rank-conj. & Yes & 46.139 & 66.395\\
Balanced Exposure & Score & Balanced & Yes & 45.567 & 65.896\\
No Common Origin & Score replay & Rank-conj. & No & 32.957 & 45.356\\
\bottomrule
\end{tabular}
\end{table}

Changing only the ordering rule shows that dynamic consensus relevance is not needed: Fixed Order exceeds Score Order by {\(0.134\%\)} on Qwen3-0.6B and {\(0.205\%\)} on Qwen3-4B. The differences are modest but consistent across scales and favor the simpler method, whose routes do not depend on update-dependent scores, atom reordering, or router hyperparameters.

The two Score-Order controls clarify the remaining mechanisms. With Score Order
held constant, rank-conjugate exposure exceeds balanced exposure by {\(0.572\%\)} and
{\(0.499\%\)} on the 0.6B and 4B models. Removing common-origin cancellation from the same routes lowers the averages by {\(13.182\%\)} and {\(21.039\%\)},
respectively, the largest degradation in this comparison. Rank-conjugate
exposure contributes a smaller but consistent gain of about {\(0.5\%\)}, while
dynamic score ordering is slightly worse than fixed order. These results favor
an algebraic origin invariant and deterministic nested coordinates without a
learned router.

\subsection{Aggregation-Rank Sensitivity}
\label{sec:compression}

The main configuration sets \(R_c=R_b\). To isolate compression, we hold
the strong client-rank profile, fixed routes, and \(R_b=64\) constant, and vary
only \(R_c\in\{8,32,64,128\}\), with an untruncated Exact control. Accuracy in
Table~\ref{tab:compression} is macro-averaged over both partitions, all three
tasks, and three seeds. Because a centered rank-\(r_i\) displacement has rank
at most \(2r_i\), the concatenated pre-compression aggregate can reach rank
\(2\sum_i r_i=320\); hence \(R_c\) can exceed \(R_b\) without changing the
anchor bank. We measure retained energy counterfactually from each recorded
Exact update:
\begin{equation}
 \rho(R_c)=1-
 \frac{\sum_{u,\ell}\|\Delta_u^\ell-
 \mathsf{Comp}_{R_c}(\Delta_u^\ell)\|_F^2}
 {\sum_{u,\ell}\|\Delta_u^\ell\|_F^2},
 \label{eq:retained_energy}
\end{equation}
where \(u\) indexes recorded updates by run and round.
This post-hoc construction applies every candidate rank to the same
Exact updates, avoiding trajectory drift in the information-loss comparison.

\begin{table}[t]
\centering
\caption{Accuracy (\%) and energy retention versus aggregation rank. Rank 64 is the
default aggregation rank for the strong-profile study; Exact performs no
truncation. Bold and underlined values denote the best and second-best accuracy
for each model.}
\label{tab:compression}
\small
\setlength{\tabcolsep}{3.3pt}
\begin{tabular}{rcccc}
\toprule
& \multicolumn{2}{c}{Qwen3-0.6B} & \multicolumn{2}{c}{Qwen3-4B}\\
\cmidrule(lr){2-3}\cmidrule(lr){4-5}
\(R_c\) & Acc. & Energy & Acc. & Energy\\
\midrule
8 & 45.92 & 72.57\% & 66.21 & 76.85\%\\
32 & \best{46.27} & 92.19\% & 66.56 & 93.41\%\\
\rowcolor{mosaiclight}
64 & \second{46.08} & 97.58\% & 66.64 & 97.95\%\\
128 & 45.65 & 99.99\% & \best{66.69} & 99.99\%\\
Exact & 45.92 & 100.00\% & \second{66.66} & 100.00\%\\
\bottomrule
\end{tabular}
\end{table}

Compression quality is monotone, but task accuracy is not. Rank 8 discards
23--27\% of update energy and is below ranks 32 and 64 on both models. Rank 64 retains about
98\% on both models and lies within {\(0.20\%\)} and {\(0.05\%\)} of the best rank on
0.6B and 4B. Increasing to rank 128 or Exact preserves virtually all remaining
energy but gives no consistent accuracy gain. Rank 64 serves as a practical default.

The rank-32 point is particularly useful operationally. In the strong profile,
whose mean client rank is 16, reducing \(R_c\) from 64 to 32 halves the
client-wise downlink ratio from \(4\times\) to \(2\times\). It still retains
92.19\% and 93.41\% of exact-update energy, obtains the best Qwen3-0.6B
accuracy, and is only {\(0.13\%\)} below the best Qwen3-4B setting. The
compression rank therefore offers a practical trade-off between communication
and accuracy: moderate
compression removes a measurable tail of update energy without necessarily
removing task-useful directions. Rank 8 is the only tested compressed setting
that falls below ranks 32 and 64 on both model scales.

\section{Discussion and Limitations}
\label{sec:limitations}

\paragraph{Interpretation.}
The results reinforce that shape compatibility and update comparability are
different requirements. \method guarantees equality at the effective-weight
and induced-function level, where LoRA factor symmetries do not matter, and
uses nested prefixes to provide explicit correspondence across ranks. Higher ranks expose more directions from the same bank. Fixed spectral ordering provides a data-independent inductive bias.

\paragraph{Scope and limitations.}
\method assumes externally assigned client ranks. It
preserves local training and uplink shape, but its common rank-\(R_c\) update increases aggregate downlink by \(2\)--\(4\times\) in our profiles; distributing an uncached atom bank would add a one-time cost. Secure aggregation and cache management therefore remain systems concerns.
Finally, our evidence covers two Qwen3 scales, three QA/reasoning datasets, ten fully participating clients, and IID or Dirichlet partitions. Generalization to other model families, partial participation, asynchrony, longer training, and more extreme rank distributions remains to be established. Calibrating cross-rank tangent-space learning strength is also complementary to the common-origin guarantee studied here.

\section{Conclusion}
\label{sec:conclusion}
\method reframes heterogeneous-rank federated LoRA as an optimization-origin problem, not only an adapter-shape problem. Model-preserving cancellation gives all clients one deterministic function-space origin, fixed rank-conjugate prefixes align their coordinates, and centered aggregation combines only learned displacements. Across Qwen3-0.6B and Qwen3-4B, \method leads 21 of 24 combinations of task, partition, and rank profile. Its absolute gains over the strongest baseline are {\(1.27\%\)} and {\(1.12\%\)}, respectively.
Fixed order also slightly outperforms dynamic scoring, while rank-64 compression retains about 98\% of update energy and remains within {\(0.20\%\)} of the best tested aggregation rank. These gains preserve local computation and uplink size at the cost of $2$--$4\times$ client-wise aggregate downlink relative to rank-specific redistribution.

\section*{Ethical Considerations}
We use public benchmarks and collect no new personal data. Federated training does not fully guarantee privacy, so deployments still require appropriate security and privacy mechanisms. Rank heterogeneity can create unequal client influence, while adapted models can inherit backbone errors and biases; both should be audited before consequential use.

\section*{GenAI Usage Disclosure}
Generative AI assisted with writing, language refinement, and grammar checks; the authors verified all technical claims, calculations, and final text.

\section*{Acknowledgments}
This work is supported by the National Key Research and Development Program of China under Grant No.~2022ZD0115903.

\bibliographystyle{ACM-Reference-Format}
\bibliography{sample-base}

\end{document}